\documentclass[10pt,twocolumn,letterpaper]{article}

\usepackage[pagenumbers]{wacv}      

\definecolor{wacvblue}{rgb}{0.21,0.49,0.74}
\usepackage[pagebackref,breaklinks,colorlinks,allcolors=wacvblue]{hyperref}
\usepackage{xcolor}
\usepackage{color}
\usepackage{xcolor,colortbl}
\usepackage{booktabs}
\usepackage{mathtools}
\usepackage{multirow}
\usepackage{tipa}
\usepackage{scalerel}
\usepackage{xspace}
\usepackage{subcaption} 

\usepackage{algorithm}
\usepackage{algpseudocode}
\usepackage{amsmath}
\usepackage{caption}
\usepackage{multicol}
\usepackage{graphicx}
\usepackage{multirow}
\usepackage{amsthm}

\theoremstyle{definition}

\theoremstyle{remark}

\definecolor{customred}{RGB}{192,0,0}
\definecolor{customyellow}{RGB}{163,124,1}
\definecolor{customgreen}{RGB}{0,176,80}
\definecolor{purple}{rgb}{0.65,0,0.65}
\definecolor{dark_green}{rgb}{0, 0.5, 0}
\definecolor{blueish}{rgb}{0.0, 0.3, .9}
\definecolor{brown}{rgb}{0.6, 0.3, 0}
\definecolor{LightCyan}{rgb}{0.88,0.95,1}

\definecolor{tabhighlight}{rgb}{0.88,0.95,1}
\definecolor{tabhighlightbluetext}{rgb}{0.2,0.4,0.8}

\definecolor{tabhighlightpurple}{rgb}{0.95,0.88,1}
\definecolor{tabhighlightpurpletext}{rgb}{0.7,0.4,0.8}

\definecolor{transparent}{cmyk}{0,0,0,0}
\definecolor{demphcolor}{RGB}{144, 144, 144}

\definecolor{ForestGreen}{RGB}{34, 139, 34}

\def\wacvPaperID{1699} 
\def\confName{WACV}
\def\confYear{2026}

\title{Low-Cost Test-Time Adaptation for Robust Video Editing}

\author{
    Jianhui Wang\textsuperscript{1*} \hspace{0.5em}
    Yinda Chen\textsuperscript{2*} \hspace{0.5em}
    Yangfan He\textsuperscript{3*} \hspace{0.5em}
    Xinyuan Song\textsuperscript{4} \hspace{0.5em}\\
    Yi Xin\textsuperscript{5} \hspace{0.5em}
    Dapeng Zhang\textsuperscript{6} \hspace{0.5em}
    Zhongwei Wan\textsuperscript{7} \hspace{0.5em}
    Bin Li\textsuperscript{8} \hspace{0.5em}
    Rongchao Zhang\textsuperscript{9} \\
    \textsuperscript{1}UESTC \quad
    \textsuperscript{2}USTC \quad 
    \textsuperscript{3}University of Minnesota--Twin Cities \quad
    \textsuperscript{4}Emory University \quad\\
    \textsuperscript{5}Nanjing University \quad
    \textsuperscript{6}Lanzhou University \quad
    \textsuperscript{7}Ohio State University, Columbus \quad \\
    \textsuperscript{8}Chinese Academy of Sciences \quad
    \textsuperscript{9}Peking University
}

\begin{document}
\maketitle
\def\thefootnote{*}\footnotetext{Equal contribution.}\def\thefootnote{\arabic{footnote}}

\begin{abstract}
Video editing is a critical component of content creation that transforms raw footage into coherent works aligned with specific visual and narrative objectives. Existing approaches face two major challenges: temporal inconsistencies due to failure in capturing complex motion patterns, and overfitting to simple prompts arising from limitations in UNet backbone architectures. While learning-based methods can enhance editing quality, they typically demand substantial computational resources and are constrained by the scarcity of high-quality annotated data. In this paper, we present Vid-TTA, a lightweight test-time adaptation framework that personalizes optimization for each test video during inference through self-supervised auxiliary tasks. Our approach incorporates a motion-aware frame reconstruction mechanism that identifies and preserves crucial movement regions, alongside a prompt perturbation and reconstruction strategy that strengthens model robustness to diverse textual descriptions. These innovations are orchestrated by a meta-learning driven dynamic loss balancing mechanism that adaptively adjusts the optimization process based on video characteristics. Extensive experiments demonstrate that Vid-TTA significantly improves video temporal consistency and mitigates prompt overfitting while maintaining low computational overhead, offering a plug-and-play performance boost for existing video editing models. Our code will be made publicly available.
\end{abstract}    
\section{Introduction}
\label{sec:intro}

\begin{figure}[t]
    \centering
    \includegraphics[width=\linewidth]{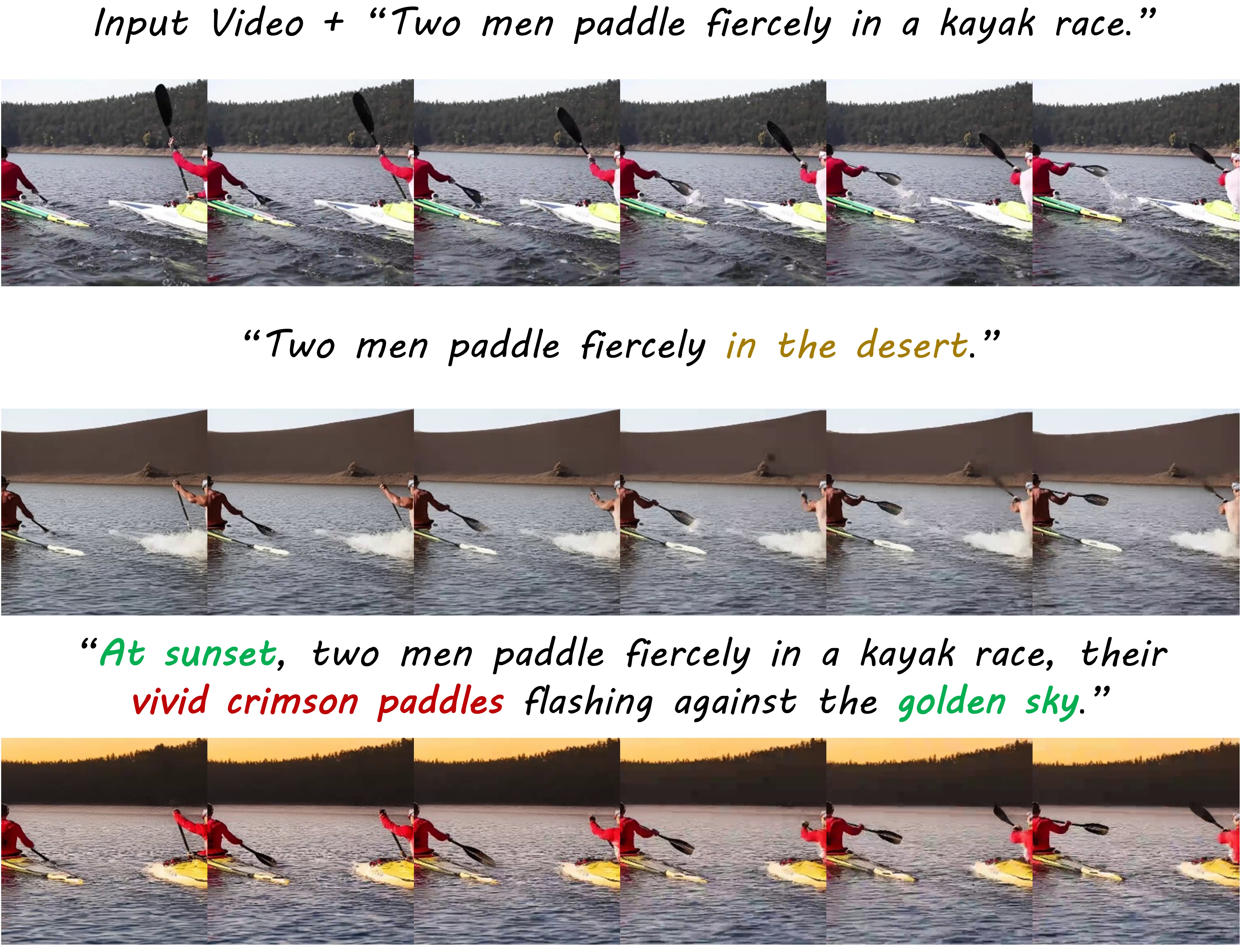}
    \caption{
    (\textbf{Top}) The input source video and prompt.
    (\textbf{Middle}) An attempt to switch the scene to a \textcolor{customyellow}{\textbf{desert}}, which clearly exposes issues in the paddle motion, including discontinuities, flickering, and occasional disappearance.
    (\textbf{Bottom}) An attempt at more detailed editing. While the ``at sunset'' and ``golden sky'' elements are \textcolor{customgreen}{\textbf{correctly}} rendered, the intended change to ``vivid crimson paddles'' was \textcolor{customred}{\textbf{not applied}}. Instead, \textcolor{customred}{\textbf{unintended modifications}} to the clothing and kayak colors occurred, reflecting a prompt overfit issue. The two cases were edited using VidToMe~\cite{li2024vidtome}.} 
    \label{fig:kayak_race}
    \vspace{-10pt}
\end{figure}

Recent years have witnessed remarkable advances in video generation, enabling the creation of impressively realistic videos from textual descriptions~\cite{ho2022imagen,hong2022cogvideo,khachatryan2023text2video,singer2022make,yan2021videogpt,yang2024cogvideox,liu2024sora}. As these technologies gain widespread adoption, video editing has emerged as an indispensable component in the content creation ecosystem, enabling creators to adjust, enhance, and modify existing video content to better align with specific visual and thematic objectives, ultimately playing a pivotal role in modern multimedia production~\cite{ceylan2023pix2video,cong2024FLATTEN,geyer2023tokenflow,li2024vidtome,qi2023fatezero,wang2023zero,wu2023tune,yang2023rerender,yang2025videograin}. These advancements have opened new horizons for multimedia production and established a foundation for further innovation in content adaptation and creative editing.

Despite the flexible solutions that existing video editing methods offer for applications ranging from style transfer to precise object replacement, two critical challenges persist. First, these approaches perform poorly in capturing complex motion patterns, particularly when handling videos with abrupt or violent movements, often resulting in temporal inconsistencies and visual artifacts. This issue stems from existing models' difficulty in effectively modeling spatiotemporal relationships, especially in scenes with dramatic motion changes. Second, due to inherent limitations in the UNet~\cite{ronneberger2015u} backbone architecture, these models frequently overfit to a limited range of training prompts, leading to suboptimal outputs when input prompts significantly deviate from the distribution in CLIP~\cite{radford2021learning} text space. This overfitting phenomenon restricts the models' ability to respond to diverse creative needs, reducing flexibility in practical applications. We present a case to illustrates both challenges in Figure~\ref{fig:kayak_race}.

To address these challenges, an intuitive approach might be to expand the training dataset with more varied motion patterns and editing prompts. However, the scarcity of training data that encompasses the full diversity of motion dynamics and editing requirements, coupled with the enormous computational resources required for training, makes this approach impractical to implement. A more practical strategy is to design a model capable of personalized adaptation for each test video, which not only reduces dependency on large-scale training data but also provides more precise editing effects.

Drawing inspiration from Test-Time Adaptation (TTA) techniques in image classification~\cite{sun2020test,xiao2024beyond}, which update model parameters for each test instance during inference, and from the effective masked reconstruction strategies demonstrated by Masked Autoencoders (MAE)~\cite{gandelsman2022test,he2021masked,tong2022videomae,wang2023videomae}, we introduce Vid-TTA, a lightweight test-time adaptation framework that personalizes optimization for each test video during inference through self-supervised auxiliary tasks. Our framework integrates a motion-aware reconstruction module that applies selective masking based on measured movement so that essential dynamic details are preserved. At the same time, it employs a text perturbation process that alters and then rebuilds input prompts, which helps the model better handle a wide range of textual descriptions. These components work together under a dynamic loss balancing strategy inspired by meta-learning. The contributions of various loss components are adjusted on the fly according to the specific characteristics of each video. Tests show that Vid-TTA improves video temporal consistency and reduces prompt overfitting while keeping computational demands low, making it a straightforward upgrade for current video editing systems.

Our main contributions can be summarized as follows:
\begin{itemize}
    \item Vid-TTA is the first framework that applies test-time adaptation to video editing, enabling the UNet backbone of existing video editing models to be dynamically fine-tuned for each test instance during inference.
    
    \item Our approach effectively addresses both temporal inconsistency and prompt overfitting through a unified self-supervised learning framework, avoiding the need for extensive labeled data or costly retraining.
    
    \item We demonstrate that Vid-TTA's low-cost and efficient framework can reliably scale to a wide range of video editing models, significantly enhancing consistency under complex motion dynamics and diverse prompt inputs, providing a practical solution for researchers and content creators alike.
\end{itemize}

\section{Related Work}
Recent progress integrates lightweight adaptation~\cite{he2025enhancing}, temporal reasoning~\cite{zhou2025glimpse,wang2025unitmge}, and multi-agent collaboration~\cite{zhou2025reagent,ye2023mplug} to address temporal incoherence and prompt overfitting. Alongside, advances in diffusion models~\cite{xin2025resurrect,xin2025lumina}, multimodal coherence~\cite{zhou2025glimpse}, and RL-enhanced generation~\cite{wang2024enhancing} further support adaptive video editing. Related efforts also tackle hallucination, alignment~\cite{zhou2023analyzing,wang2023evaluation,zhou2024calibrated,zhou2025anyprefer}, and scene-level generation~\cite{yang2024wcdt,he2024ddpm}.
\subsection{Video Editing}
Existing video editing methods generally leverage pre-trained diffusion models to edit videos by refining the frame generation process for semantic changes. Early works such as Tune-A-Video~\cite{wu2023tune} employ one-shot tuning, where only key spatiotemporal components are updated to preserve original motion while incorporating new content through text prompts. Pix2Video~\cite{ceylan2023pix2video} and vid2vid-zero~\cite{vid2vid-zero} introduce feature propagation and null-text inversion mechanisms to maintain structural fidelity without extensive re‑training, while DMT~\cite{yatim2023spacetime} introduces a novel motion transfer loss that enables cross-category motion editing while preserving the original motion patterns. Render-A-Video~\cite{yang2023rerender} introduced a two-stage framework that first edits selected key frames using optical flow guidance and then propagates these edits to the remaining frames. TokenFlow~\cite{geyer2023tokenflow} exploits diffusion feature consistency to achieve robust video editing without additional fine-tuning. 

More recently, FLATTEN~\cite{cong2024FLATTEN} and FateZero~\cite{10378281}, further tackle inter‑frame consistency challenges by incorporating optical flow-based attention modulation and token merging strategies. Additionally, approaches like Ground-A-Video~\cite{jeong2024groundavideo} and ControlVideo~\cite{zhang2024controlvideo} integrate extra control signals, such as edge maps or depth cues, to enable finer-grained editing while ensuring temporal coherence. VidToMe~\cite{li2024vidtome} and VideoGrain~\cite{yang2025videograin} optimize the internal attention mechanisms to better align features across frames, thereby minimizing inconsistencies.

\subsection{Test-Time Adaptation}
Test-Time Adaptation (TTA) is a rapidly emerging paradigm that enables models to adjust to target domains during inference without requiring retraining on large-scale datasets~\cite{xiao2024modeladaptationtesttime}. Various approaches have been developed, primarily leveraging self-supervised auxiliary tasks to improve adaptation in fields such as image processing~\cite{bahmani2023semanticselfadaptationenhancinggeneralization, yeo2023rapidnetworkadaptationlearning}, video analysis~\cite{lin2023video, 10655110, yi2023temporal}, and 3D recognition~\cite{shim2024cloudfixer,shin2022mm,wang2024backpropagation}. Most studies utilizing TTA focus on developing self-supervised auxiliary tasks and integrating them with the main task to enhance adaptation.

\subsection{Masked Language and Vision Modeling}
Masked modeling has emerged as a powerful paradigm for both text and video domains. In the text domain, the seminal masked language modeling (MLM) framework introduced by BERT~\cite{Devlin2019BERTPO} has evolved significantly, with work such as DiffusionBERT~\cite{He2022DiffusionBERTIG} demonstrating that integrating diffusion processes can enhance generative and contextual understanding. Further architectural refinements, such as NextLevelBERT~\cite{czinczoll-etal-2024-nextlevelbert}, have improved long-document representation. In parallel, masked video modeling has progressed from early methods like VideoMAE~\cite{tong2022videomae} to more recent advances such as VideoMAE V2~\cite{wang2023videomaev2}, which employs a dual masking strategy to reduce computational cost, and M³Video~\cite{kwon2023masked} that leverages optical flow-guided masked motion prediction to better capture dynamic patterns. These innovations demonstrate that the “mask then predict” approach is not only scalable but also adaptable, leading to more robust representations for both textual and spatiotemporal information.

\section{Method}
We begin by identifying two major challenges in video editing and introducing our approach in Section~\ref{sec:overview}. Section~\ref{sec:spatiotemporal} describes our motion-aware spatiotemporal representation learning mechanism. Next, Section~\ref{sec:semantic} presents our prompt augmentation strategy and reconstruction task to address prompt overfitting. Finally, Section~\ref{sec:dynamic} details our adaptive loss balancing framework.

\begin{figure*}[htbp]
    \centering
    \includegraphics[width=\linewidth]{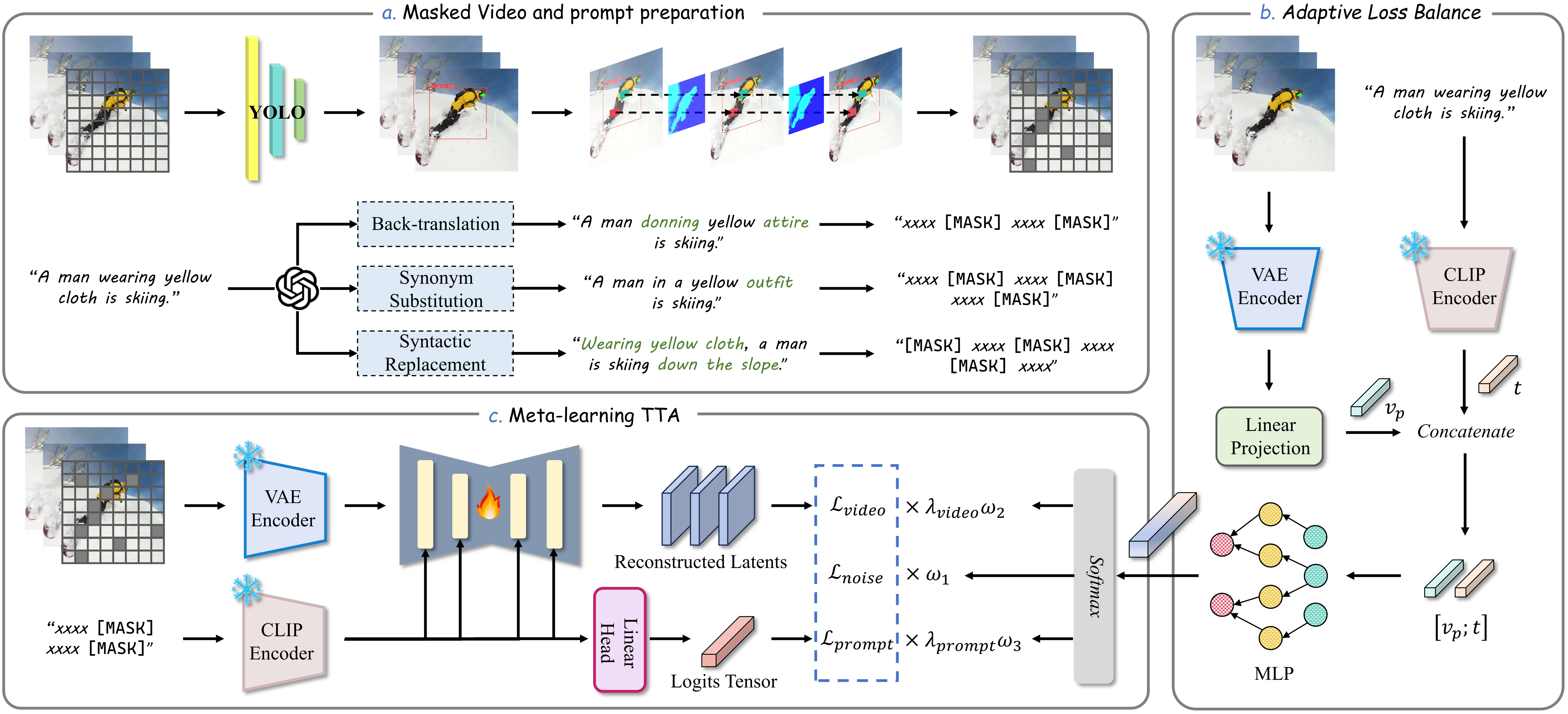}
    \caption{Overview of Vid-TTA. 
    (a) We first detect moving objects with YOLO and apply optical flow estimation inside the bounding box to identify high-motion patches, creating masked video segments. In parallel, we perform prompt augmentation (back-translation, synonym substitution, and syntactic reformation) on the editing prompt, followed by a token-level masking procedure. 
    (b) The original video and prompt are passed through a VAE encoder and a text encoder, respectively. Three losses (noise prediction, video reconstruction, and prompt reconstruction) are computed. These losses are weighted and balanced by a multi-scale feature-based module that concatenates latent representations from the video and text encoders, using MLP to generate dynamic weights.
    (c) During fine-tuning, we update UNet parameters using auxiliary reconstruction tasks on both masked video and prompt inputs, with dynamic loss weighting guiding the update.
}
    \label{fig:overview}
\end{figure*}

\subsection{Unified Framework Overview}
\label{sec:overview}

Given an input video $V$ and an editing prompt $T$, the goal of video editing is to generate a new video $V^*$ that preserves the temporal consistency of $V$ while incorporating the semantic modifications specified by $T$. In practice, two fundamental challenges arise. First, $V^*$ must maintain the complex motion dynamics of $V$, particularly when handling rapid or intricate movements. Second, video editing models typically exhibit overfitting to simple prompts, constraining their capacity to generalize across diverse textual descriptions. This overfitting phenomenon, commonly observed in both text-to-image and text-to-video models, stems primarily from the UNet backbone being trained with limited prompt diversity. Consequently, when presented with novel or more nuanced prompts at inference time, the model struggles to capture the intended semantic variations.

To address these challenges, we propose a test-time adaptation framework illustrated in Figure~\ref{fig:overview}. Our approach incorporates two complementary tasks: (i) employing masked autoencoding in detected motion regions to guide the reconstruction of video latent representations, thereby enforcing motion consistency; and (ii) implementing a prompt augmentation and reconstruction mechanism to mitigate prompt overfitting. Mathematically, our approach can be formulated as:
\begin{equation}
\theta^* = \arg\min_\theta \mathcal{L}_{total}(\theta; V, T),
\end{equation}
where $\theta$ denotes the model parameters and $\mathcal{L}_{total}$ represents the total loss function. By simultaneously enhancing temporal consistency and semantic fidelity, our method produces videos with smooth motion patterns and precise edits.

\subsection{Spatio-temporal Representation Learning}
\label{sec:spatiotemporal}

To effectively capture and preserve motion dynamics in videos, we introduce an adaptive spatio-temporal representation learning approach that focuses on key motion regions. Our method begins with object detection to localize moving entities, followed by optical flow analysis to identify significant motion areas.

Specifically, we apply YOLOv8~\cite{redmon2016lookonceunifiedrealtime} to perform object detection on each frame $f_t$ of the video, producing a bounding box $B_t = (x_t, y_t, w_t, h_t)$. Given a video tensor $v \in \mathbb{R}^{T \times C \times H \times W}$, we partition each frame into patches of size $p_h \times p_w$. The total number of patches per frame is:
\begin{equation}
N_p = (H/p_h) \times (W/p_w).
\end{equation}
Within the bounding box $B_t$, the number of patches covering the object is:
\begin{equation}
N_{B_t} = (h_t/p_h) \times (w_t/p_w).
\end{equation}
We define mask ratios $r_f$ and $r_b$ for the object and background regions, respectively. The number of patches to be masked is:
\begin{equation}
M_f = r_f \cdot N_{B_t}, \quad M_b = r_b \cdot (N_p - N_{B_t}).
\end{equation}

Subsequently, we utilize GMFlow~\cite{xu2022gmflow} to compute the optical flow $F_t$ between consecutive frames and derive a motion intensity $m_t(i)$ for each patch. Patches within $B_t$ are ranked by their motion intensity, with the top $M_f$ patches selected for masking, while $M_b$ background patches are randomly masked. The resulting mask $M_t$ is applied to the latent representation $z_t$ from the VAE encoder: $\tilde{z}_t = M_t \odot z_t$, where $\odot$ denotes element-wise multiplication. The video reconstruction loss is:
\begin{equation}
\mathcal{L}_{video} = \frac{1}{N} \sum_{i=1}^{N} \lVert z_t(i) - \tilde{z}_t(i) \rVert_2^2.
\end{equation}
This targeted approach enables the model to focus on the most challenging dynamic elements of the video, thereby enhancing motion consistency without incurring additional computational overhead.

\subsection{Semantic Representation Space Exploration}
\label{sec:semantic}

To address prompt overfitting, we develop a semantic representation space exploration framework that enhances model generalization through systematic prompt perturbation. Given an original editing prompt $T$, we generate a set of augmented prompts $\{T_a^i\}_{i=1}^n$ that preserve the semantic content while exploring different regions of the text representation space.

Our augmentation strategy employs three primary techniques: back-translation, synonym replacement, and syntactic reformation. These techniques produce semantically equivalent prompts that differ in expression, enabling the model to learn more robust mappings in the text representation space.

Subsequently, we implement a masked prompt reconstruction task to facilitate effective learning. We apply a random masking strategy to the augmented prompts, replacing selected tokens with \texttt{[MASK]} tokens to generate masked prompts $T_m$. These masked prompts are processed through the text encoder to produce hidden representations, which are then used by a linear reconstruction head $\phi(\cdot)$ to predict the original token distribution. The prompt reconstruction loss is:
\begin{equation}
\mathcal{L}_{prompt} = \text{CE}(\phi(T_m), T_a).
\end{equation}
This self-supervised approach compels the model to learn more robust text representations capable of handling various textual variations, thereby mitigating prompt overfitting.

\begin{figure*}[t]
    \centering
    \includegraphics[width=0.85\textwidth]{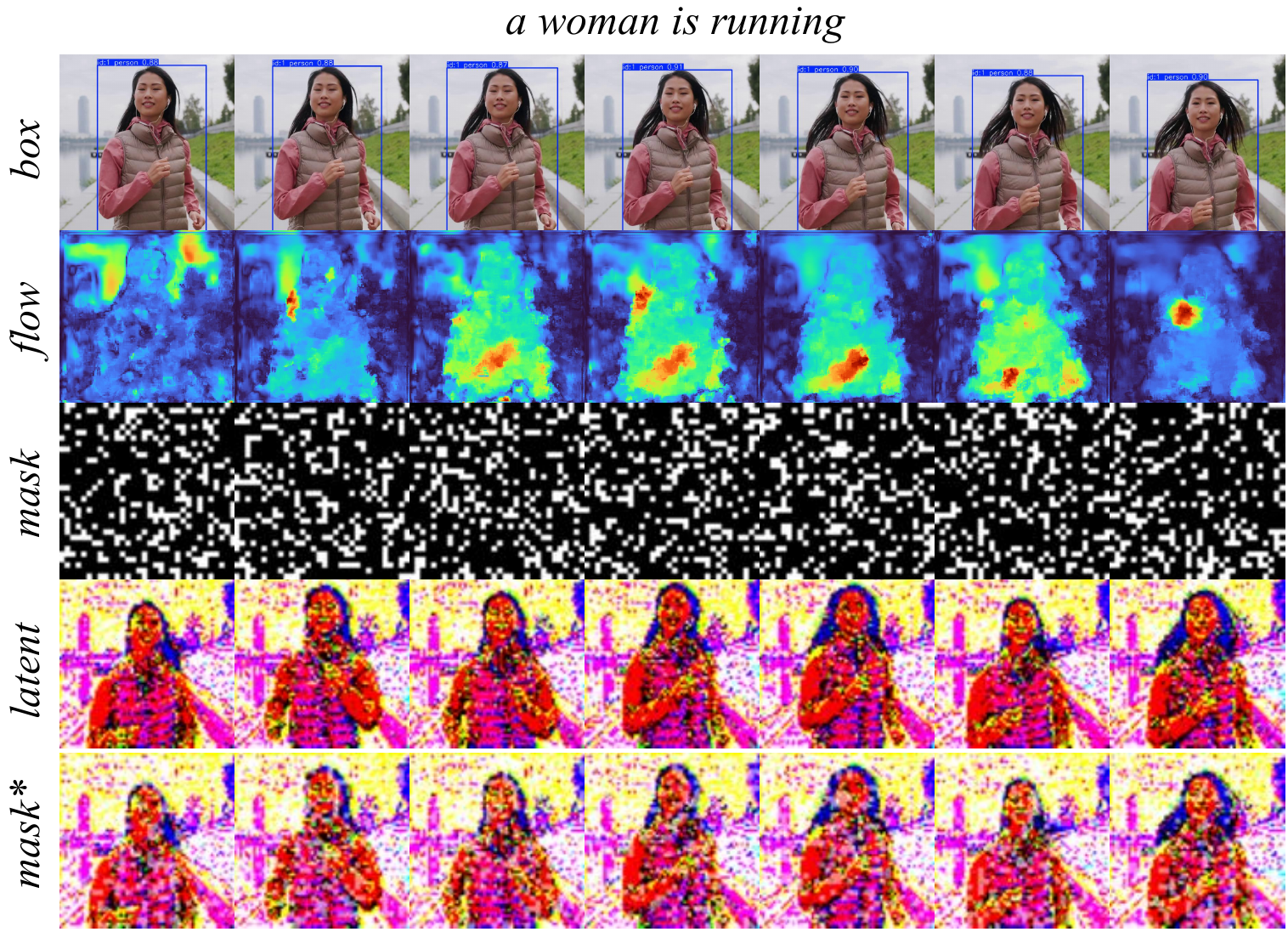}
    \caption{In the top row (box), YOLOv8 detects and draws bounding boxes around the moving subject in each video frame; the second row (flow) shows GMFlow‑computed optical‑flow intensity heatmaps that highlight regions of high motion; the third row (mask) displays the binary masks generated by selecting the top motion patches inside the detected box and random patches in the background; the fourth row (latent) depicts the original VAE‑encoded latent representations of each frame; and the bottom row (mask*) presents the reconstructed latents after masked autoencoding, demonstrating how the model fills in masked dynamic regions to reinforce motion consistency.}
    \label{fig:visual}
\end{figure*}

\subsection{Meta-Learning TTA}
\label{sec:dynamic}
Our goal is to leverage self-supervised auxiliary tasks to adjust the UNet weights during inference. To maximize the effectiveness of test-time adaptation, we introduce a meta-learning driven adaptive loss balancing mechanism that automatically adjusts the relative importance of different loss components based on instance-specific features. We extract a global video feature $v$ from the latent representation $z_t$ through spatial averaging:
\begin{equation}
v = \frac{1}{HW} \sum_{i=1}^{H} \sum_{j=1}^{W} z_t(i,j) \in \mathbb{R}^C.
\end{equation}
We project this video feature into the text encoder's hidden space via a linear mapping $g(\cdot)$:

\begin{equation}
v_p = g(v) \in \mathbb{R}^d.
\end{equation}
Simultaneously, we derive a global text feature from the hidden states $h \in \mathbb{R}^{L \times d}$ of the text encoder:

\begin{equation}
t = \frac{1}{L} \sum_{k=1}^{L} h_k \in \mathbb{R}^d.
\end{equation}
These features are concatenated and processed through a multilayer perceptron (MLP) to generate a three-dimensional weight vector:

\begin{equation}
\text{softmax}(\psi([v_p;t])) = [w_1, w_2, w_3] \in \mathbb{R}^3.
\end{equation}
The total loss function is defined as:

\begin{equation}
\mathcal{L}_{total} = w_1 \mathcal{L}_{noise} + w_2 \lambda_{video} \mathcal{L}_{video} + w_3 \lambda_{text} \mathcal{L}_{prompt},
\end{equation}
where $\lambda_{video}$ and $\lambda_{text}$ are scaling factors that normalize the different loss components. This adaptive balancing mechanism allows the model to dynamically adjust its learning strategy according to the specific characteristics of each video, resulting in more effective test-time adaptation.

\begin{figure*}[t]
    \centering
    \includegraphics[width=0.8\textwidth]{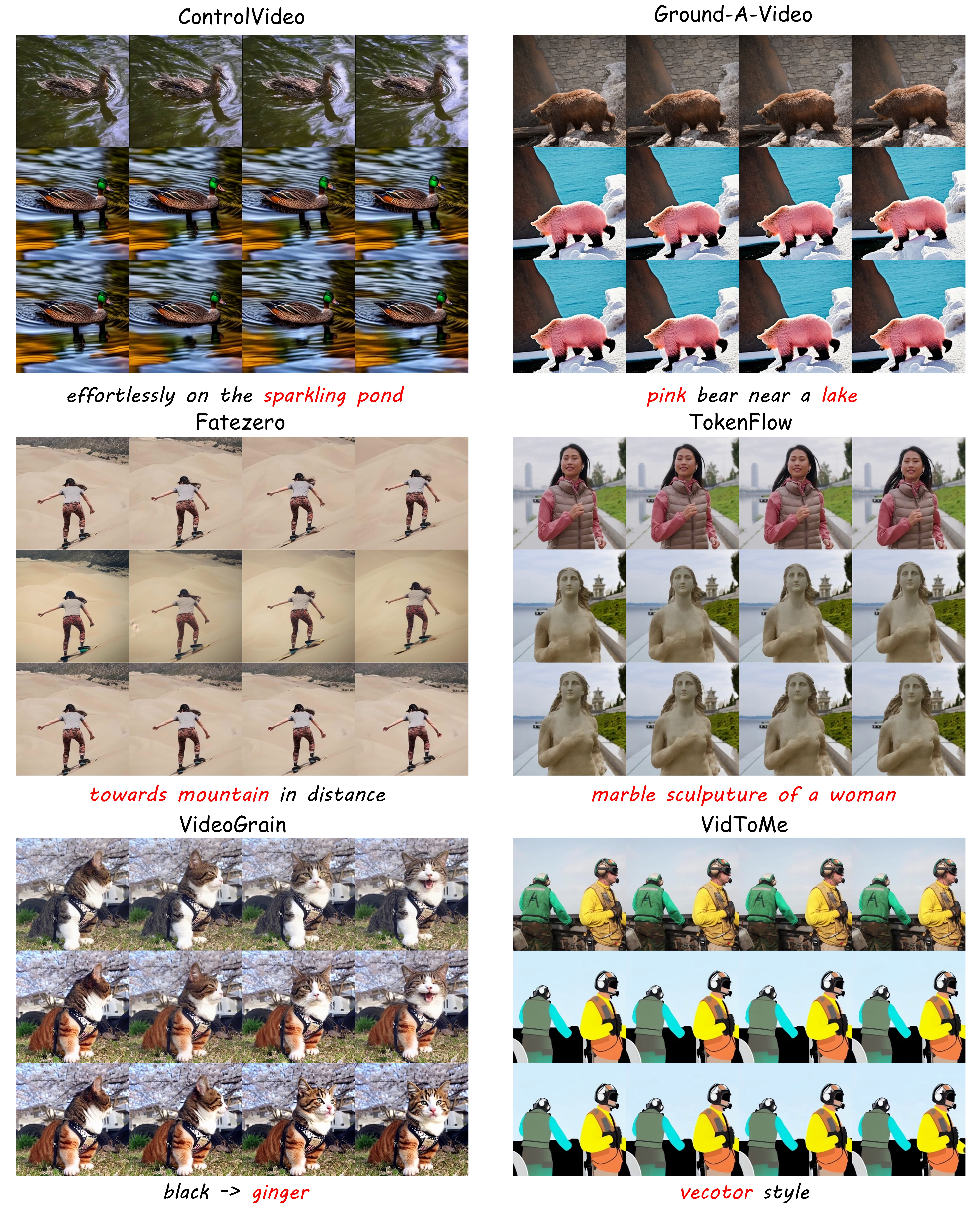}
    \caption{Qualitative comparison of six video editing models before and after applying Vid-TTA. 
    From top to bottom in each sub-figure: the original input frames, the baseline results, and the baseline enhanced by Vid-TTA. 
    Visually, our method preserves motion details while achieving more accurate and diverse edits.}
    \label{fig:compare_visual}
    \vspace{-10pt}
\end{figure*}
\section{Experiments}
\subsection{Experimental Settings}
We conducted experiments on a wide range of video editing algorithms to validate the effectiveness and robustness of our proposed method. For experiments using a 3D UNet backbone, we performed tuning on an NVIDIA A100 GPU with 80GB of memory, whereas experiments with a 2D UNet backbone were carried out on a single NVIDIA 4090 GPU. The masking ratios are set to 0.75 for object regions and 0.2 for background regions. Each original prompt is augmented to yield 3 additional prompts, and 30\% of tokens in these prompts are randomly masked. Both scaling factors \(\lambda_{\text{video}}\) and \(\lambda_{\text{text}}\) are set to 0.1.

\begin{figure*}[t]
    \centering
    \includegraphics[width=\linewidth]{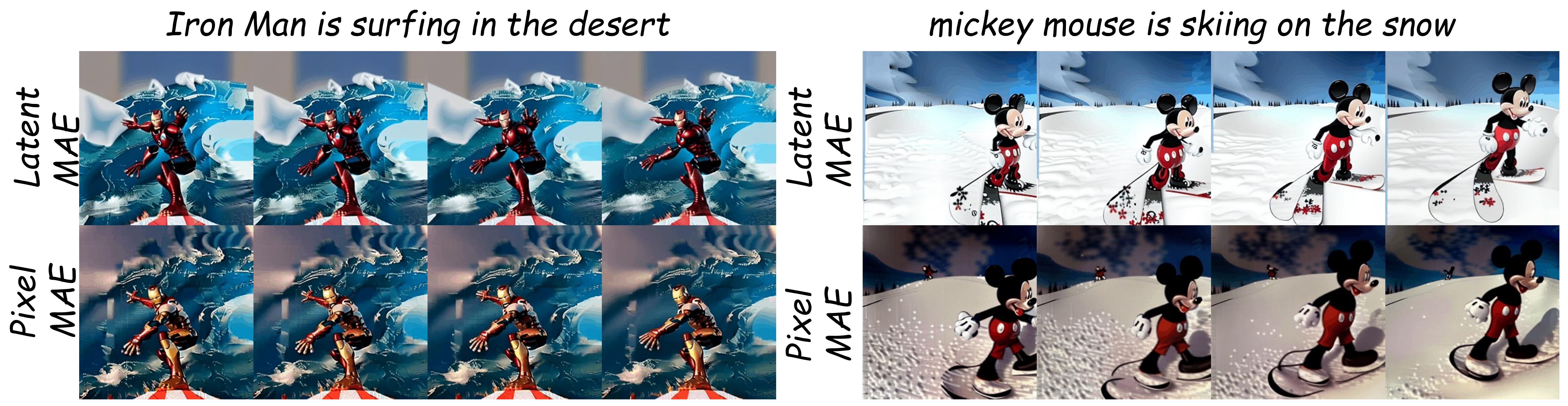}
    \caption{Comparison of Latent MAE and Pixel MAE effects on video quality: In scenes like 'Iron Man surfing in the desert' and 'Mickey Mouse skiing on the snow'}
    \label{fig:ablation_study_mae}
    \vspace{-10pt}
\end{figure*}

\section{Visualization Results}
In Figure~\ref{fig:visual}, we visualize the key stages of our test‑time adaptation (TTA) pipeline on a representative clip prompted with “a woman is running.” This visualization confirms that our motion‐aware masking and reconstruction mechanism effectively guides the model to focus on—and faithfully reproduce—critical motion details during video editing. The frames validate that our spatio‑temporal representation learning module indeed forces the model to concentrate on—and faithfully reproduce—complex motion patterns during video editing.

Figure~\ref{fig:compare_visual} compares six representative video editing methods, showing their outputs before and after applying Vid-TTA. From top to bottom in each sub-figure: the original input frames, the baseline results, and the baseline enhanced by Vid-TTA.

\subsection{Ablation Study}

\noindent\textbf{Ablation Visualizations.} 
Figure~\ref{fig:ablation_study_mae} compares using a latent‐space MAE loss versus a pixel‐space MAE loss on two challenging editing scenarios (“Iron Man surfing in the desert” and “Mickey Mouse skiing on the snow”). We observe that latent MAE produces much smoother temporal transitions and preserves fine-grained object details, whereas pixel MAE often introduces flickering and color jitter.  

\begin{figure}[htbp]
    \centering
    \includegraphics[width=\linewidth]{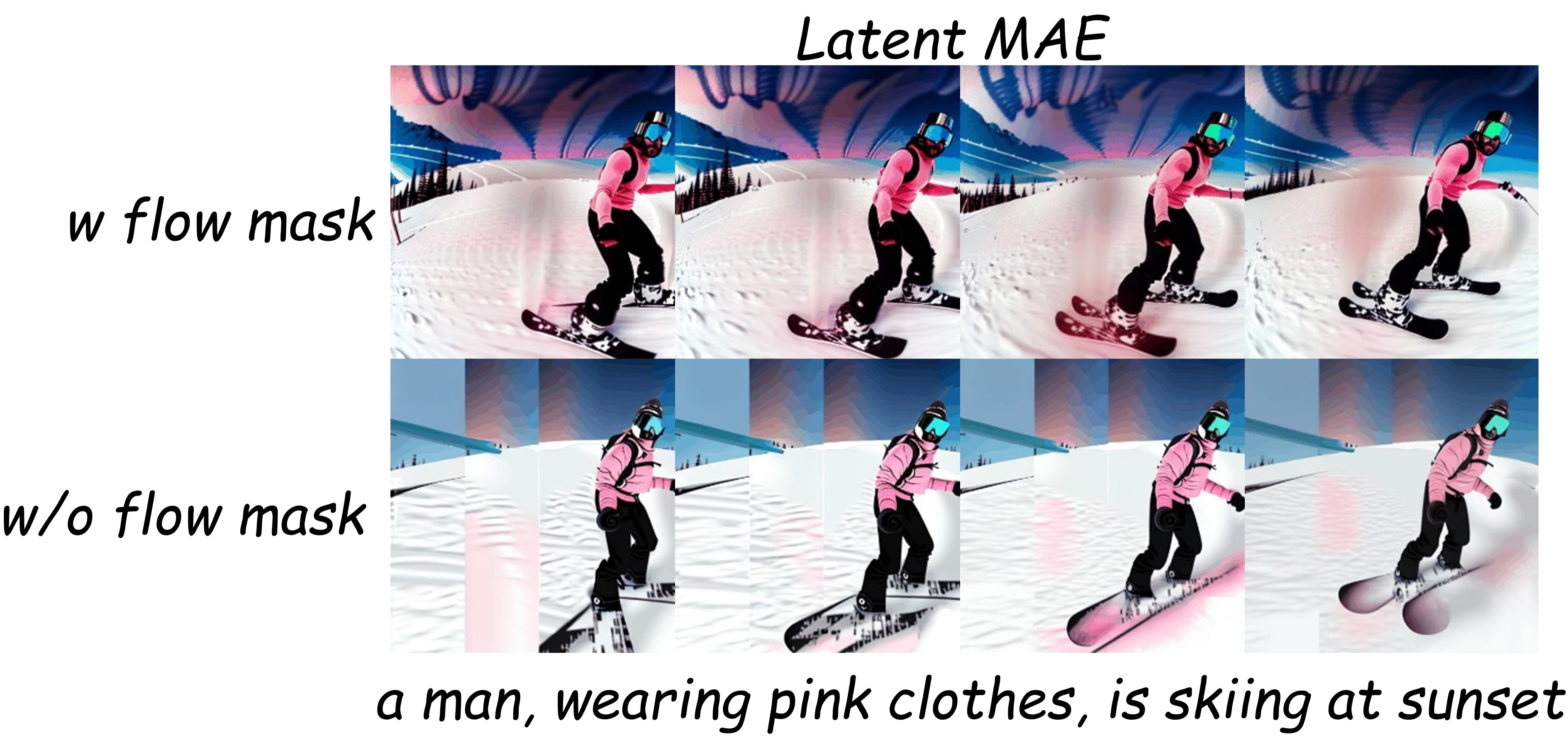}
    \caption{Comparison of Latent MAE with and without flow mask}
    \label{fig:ablation_study_flow_mask}
    \vspace{-10pt}
\end{figure}

Figure~\ref{fig:ablation_study_flow_mask} further isolates the impact of our motion‐aware masking: applying the flow‐based mask during latent MAE (top) reduces motion artifacts—especially around fast‐moving edges—compared to vanilla latent MAE without any masking (bottom). These visual results corroborate our quantitative findings, demonstrating that both the choice of reconstruction space and the motion‐guided mask are critical for high‐quality video editing.

\section{Conclusion}

In this work we present Vid‑TTA, a lightweight plug and play test time adaptation framework that tackles both temporal inconsistencies and prompt overfitting in video editing. By combining motion aware masked reconstruction with systematic prompt augmentation and coordinating these through a meta learned instance specific loss weighting mechanism, Vid‑TTA tailors its updates to each input video without costly retraining or extra labels. Extensive experiments on leading editing models demonstrate clear gains in motion coherence, semantic fidelity and perceptual quality as confirmed by human studies. Vid‑TTA offers a practical path toward more robust and versatile video editing systems for researchers and content creators alike.

{
    \small
    \bibliographystyle{ieeenat_fullname}
    \bibliography{main}
}

\end{document}